\documentclass[conference]{IEEEtran}
\IEEEoverridecommandlockouts
\usepackage{cite}
\usepackage{amsmath,amssymb,amsfonts}
\usepackage{algorithmic}
\usepackage{graphicx}
\usepackage{textcomp}
\usepackage{xcolor}
\usepackage{caption, subfig}
\usepackage{amsmath}
\usepackage{amssymb}
\usepackage{cases}
\usepackage{multirow}
\usepackage{multicol}
\usepackage{adjustbox}
\usepackage{makecell}
\usepackage{url}
\usepackage{enumitem}

\usepackage{diagbox} 
\def\BibTeX{{\rm B\kern-.05em{\sc i\kern-.025em b}\kern-.08em
    T\kern-.1667em\lower.7ex\hbox{E}\kern-.125emX}}

\begin{document}

\title{LoFi: Vision-Aided Label Generator for Wi-Fi Localization and Tracking}


\author{%
    \IEEEauthorblockN{Zijian Zhao\IEEEauthorrefmark{1}\IEEEauthorrefmark{2}, Tingwei Chen\IEEEauthorrefmark{1}, Fanyi Meng\IEEEauthorrefmark{1}\IEEEauthorrefmark{3}, Zhijie Cai\IEEEauthorrefmark{1}\IEEEauthorrefmark{3}, Hang Li\IEEEauthorrefmark{1}, Xiaoyang Li\IEEEauthorrefmark{1}, Guangxu Zhu\IEEEauthorrefmark{1}}%
    \IEEEauthorblockA{\IEEEauthorrefmark{1} Shenzhen Research Institute of Big Data \\ %
    \IEEEauthorrefmark{2} School of Computer Science and Engineering, Sun Yat-sen University \\ %
    \IEEEauthorrefmark{3} School of Science and Engineering, The Chinese University of Hong Kong, Shenzhen\\ %
    Email:\{zhaozj28\}@mail2.sysu.edu.cn,\{tingweichen,fanyimeng,zhijiecai\}@link.cuhk.edu.cn, \\
    \{hangdavidli,lixiaoyang,gxzhu\}@sribd.cn
    }%
   
}

\maketitle

\begin{abstract}
Data-driven Wi-Fi localization and tracking have shown great promise due to their lower reliance on specialized hardware compared to model-based methods. However, most existing data collection techniques provide only coarse-grained ground truth or a limited number of labeled points, significantly hindering the advancement of data-driven approaches. While systems like LiDar can deliver precise ground truth, their high costs make them inaccessible to many users. To address these challenges, we propose LoFi, a vision-aided label generator for Wi-Fi localization and tracking. LoFi can generate ground truth position coordinates solely from 2D images, offering high precision, low cost, and ease of use. Utilizing our method, we have compiled a Wi-Fi tracking and localization dataset using the ESP32-S3 and a webcam. The code and dataset of this paper is available at \url{https://github.com/RS2002/LoFi}.
\end{abstract}

\begin{IEEEkeywords}
Wi-Fi Sensing, Wi-Fi Tracking, Wi-Fi Localization, Multi-Modal, Channel Statement Information (CSI)
\end{IEEEkeywords}

\section{Introduction}

Indoor localization and tracking tasks have long been a challenge. The most common method, GPS, often fails to work reliably indoors, as the signals are easily blocked, distorted, and weakened by building materials and structures, resulting in poor satellite visibility and geometry for accurate positioning. 
Vision \cite{indoor_location_vision}, LiDar \cite{indoor_location_lidar}, and bluetooth \cite{indoor_location_bluetooth} are some common indoor localization methods, but they have their own shortcomings, like high requirements for line of sight in the case of vision, high cost of LiDar, and low accuracy of bluetooth.

In contrast, Wi-Fi sensing can efficiently address these problems. When people move within a Wi-Fi-enabled indoor environment, their presence and movements can alter the propagation of Wi-Fi signals, causing changes in the Received Signal Strength Information (RSSI) and Channel State Information (CSI) at nearby Wi-Fi access points or client devices. By analyzing these changes in Wi-Fi signal characteristics, it is possible to infer the location, movement, and even activities of the person within the indoor space. Wi-Fi sensing has been viewed as a very suitable technology for indoor localization and tracking, as it has shown promising performance in many works \cite{SiFi}. Additionally, since Wi-Fi devices are ubiquitous in many indoor scenarios, these methods can be realized without the need for any extra devices.

Wi-Fi localization and tracking methods can be broadly classified into two categories: model-based methods and data-driven methods. Model-based methods typically require multiple receivers or antennas as they rely on leveraging spatial diversity and multi-path analysis \cite{wu2021witraj}. However, this requirement may not always be practical, especially in home scenarios where a single Wi-Fi router is commonly used. In contrast, data-driven methods do not have such limitations, but they often require a large amount of training data. Currently, most Wi-Fi localization and tracking datasets rely on ground truth data obtained through manual tagging or LiDar-based methods. Manual tagging can only provide coarse-grained position information or a limited number of data points, which can adversely impact the model's generalization capacity. Additionally, manual tagging is time-consuming and labor-intensive. While LiDar-based methods can provide accurate results, their high cost makes them inaccessible to many users. As a result, most current data-driven methods are limited to classification tasks that can only identify a coarse-grained region where a person is located or the general trajectory shape of person's movements \cite{SiFi,werner2014reliable}, rather than precise tracking capabilities .

To address this issue, we propose LoFi, an easy, quick, and low-cost method for collecting Wi-Fi tracking and localization datasets by utilizing vision to extract ground truth locations of individual. This approach employs Object Detection (OD) techniques to capture the coordinates of single person in pixel space and convert it to physical space. The main contributions of this work are: 
\begin{itemize}
    \item  We propose LoFi, a vision-aided label generator for \underline{Lo}calization and tracking tasks in Wi-\underline{Fi} sensing. Our method enables rapid and cost-effective dataset collection without any specific requirement for camera, Wi-Fi devices, or predefined movement trajectories.

    \item  We present a Wi-Fi tracking dataset using our LoFi method, including CSI, RSSI, timestamp, coordinates, and person ID. It is suitable for various tasks, including tracking, localization, person identification. Additionally, we provide benchmark methods to demonstrate the efficiency of LoFi.
\end{itemize}


\begin{table*}[t!]
\caption{Dataset Description: `--' represents information not mentioned in the paper and `*' represents the dataset is publicly available.}
    \centering
    \begin{adjustbox}{width=1.0\textwidth}
        \begin{tabular}{|c|c|c|c|c|c|c|c|c|}
        \hline
        \multirow{2}{*}{\textbf{Dataset}} & \multicolumn{2}{c|}{\textbf{Transmitter}} & \multicolumn{2}{c|}{\textbf{Receiver}} & \multirow{2}{*}{\textbf{Scale}} & \multirow{2}{*}{\textbf{Modality}} & \multirow{2}{*}{\textbf{Device}} & \multirow{2}{*}{\textbf{\makecell{Label Generation \\ Method}}}\\
        \cline{2-5}
        & \textbf{Site} & \textbf{Antennas} & \textbf{site} & \textbf{antennas} & & & & \\
          \hline
         \cite{DeepFi,PhaseFi,SiFi} & 1 & 1 & 1 & 3 & 56,040-300,000 frames & CSI & Intel WiFi Link 5300 NIC & \multirow{8}{*}{\makecell{Assign \\ Previously}} \\
        \cline{1-8}
         \cite{zhu2021bls}*,\cite{zhu2022intelligent}*,\cite{sanam2018improved,sanam2020multi,jing2019learning} & 1 & 3 & 1 & 3 & 6,000-504,000 frames & CSI & Intel WiFi Link 5300 NIC &  \\
        \cline{1-8}
         \cite{yan2021extreme} & 1 & -- & 1 & 3 & -- & CSI \& RSSI & Intel WiFi Link 5300 NIC &  \\
        \cline{1-8}
          \cite{roy2019juindoorloc}* & 172 & -- & 4 & -- & 25,364 frames & RSSI & mobile phone &   \\
        \cline{1-8}
        \cite{mendoza2018long}* & 620 & -- & 1 & -- & 3,696 frames & RSSI & mobile phone &   \\
        \cline{1-8}
         \cite{li2022wivelo}* & 1 & -- & 2 & 3 & 216 instances & CSI & Intel WiFi Link 5300 NIC &  \\
        \cline{1-8}
       \cite{wu2021witraj}* & 1 & 3 & 3 & 3 & 60 instances & CSI & Intel WiFi Link 5300 NIC \& Camera & \\
        \cline{1-8}
        \cite{arnold2019novel}* & 1 & 1 & 1 & 16 & 17,000 frames & CSI &  \makecell{USRP \& SDR-equipped \\ Vacuum-cleaner Robot} &   \\
        \hline \hline
        Ours* & 1 & 1 & 1 & 1 & 210,000 frames & CSI \& RSSI & ESP32-S3 \& Webcam & Vision-Aided Method \\
        \hline
        \end{tabular}
        \end{adjustbox}
\label{tab:dataset}
\end{table*}

\section{Related Work}


In traditional Wi-Fi localization datasets, most approaches assign predetermined positions for individuals to stand at while collecting data \cite{PhaseFi}. This method only captures data at specific points, significantly compromising the generalization capability of data-driven techniques. Additionally, some localization datasets divide the room into regions, providing only the region in which individuals are located, which restricts them to coarse-grained localization \cite{SiFi}.

For Wi-Fi tracking datasets, most predefined trajectories in advance, limiting their use to training models that can only identify the shape of those trajectories \cite{werner2014reliable}. Some datasets mark specific points on the ground and assume individuals move uniformly between consecutive points. In this scenario, positions can be calculated using interpolation methods based on timestamps \cite{zhang2021deep}. However, the assumption of uniform motion can lead to inaccuracies in the ground truth data. Although some studies utilize cameras for accurate interpolation, the marked points still constrain individuals' movement \cite{wu2021witraj}, which negatively impacts model generalization.

In contrast, our LoFi method captures individuals' motion without restrictions by using cameras to generate their coordinates, avoiding the limitations of predetermined positions or trajectories. A detailed comparison is presented in Table \ref{tab:dataset}.

\section{Methodology}

\begin{figure*}[t!]
\centering 
\includegraphics[width=\textwidth]{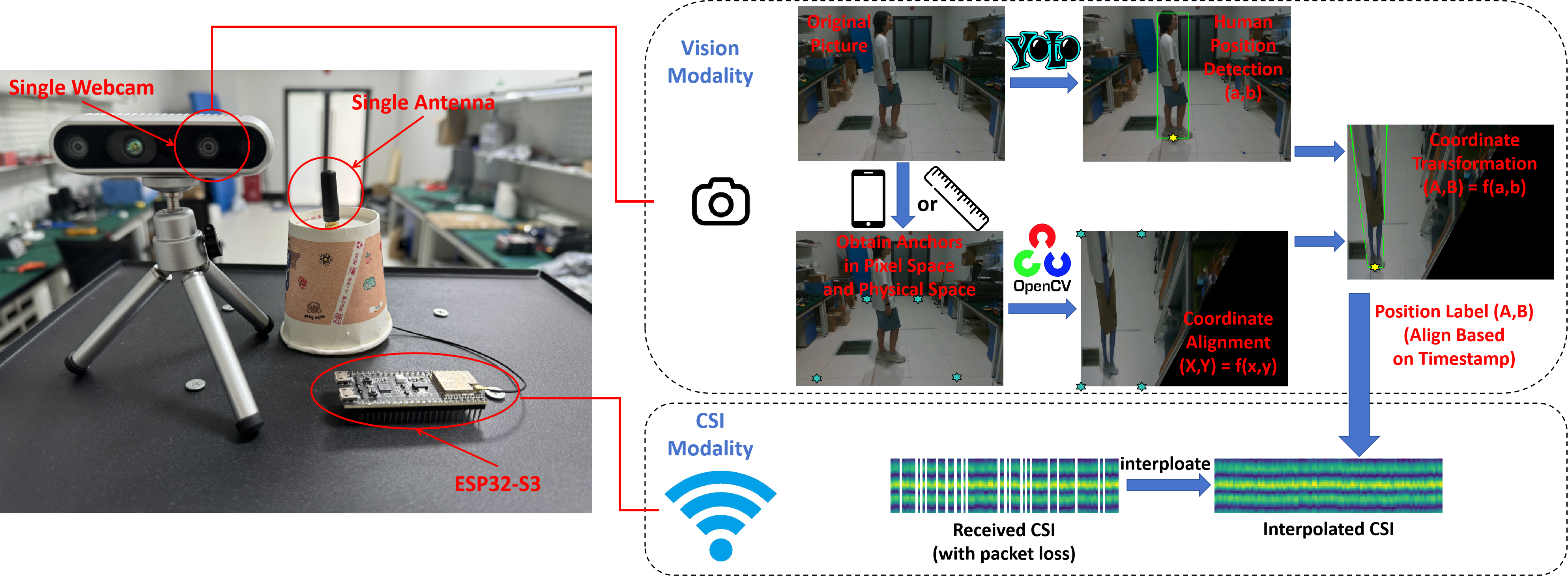}
\caption{Workflow: \textbf{Step 1}: Collect vision modality and CSI modality data simultaneously. \textbf{Step 2}: Detect individual position in the image based on OD methods. \textbf{Step 3}: Obtain the coordinates of four non-collinear anchors and construct the mapping relationship between pixel space and physical space. \textbf{Step 4}: Transfer the individual position from pixel space to physical space. \textbf{Step 5}: Use interpolation methods to fill in the lost CSI packets. \textbf{Step 6}: Align the CSI modality and vision modality to match the individual position to each CSI packet.}
\label{fig:workflow}
\end{figure*}

In this paper, we propose a system for collecting datasets for network training in Wi-Fi localization and tracking tasks. Currently, most neural networks in this domain take original or processed signals, such as CSI and RSSI, as inputs, and output either fine-grained person coordinates or coarse-grained person positions (i.e., dividing the space into multiple regions and classifying the person's location within a specific region). Based on this framework, our system workflow is illustrated in Fig. \ref{fig:workflow}. In this setup, a camera collects images that enable the extraction of real-world person coordinates, while a Wi-Fi receiver (RX) gathers signal data that will be aligned with the corresponding coordinate labels.

\subsection{Vision-aided Label Generation}
To obtain real-world coordinates of individuals, we first capture their locations in 2D images and then convert these coordinates from pixel space to physical space. In our system, the camera captures images at a frequency of $f_1$. We utilize a trained OD model to identify the locations of individuals. Most OD models produce three outputs: anchor boxes $a = [a_1, a_2, \ldots, a_n]$, object categories $c = [c_1, c_2, \ldots, c_n]$, and the confidence levels of each detection result $p = [p_1, p_2, \ldots, p_n]$. Due to potential false detections, we select the anchor box with the highest confidence in the `person' category as the estimated person location:

\begin{equation}
\hat{a} = a[\arg\max \mathbf{1}\{y = \text{`person'}\} \cdot p] \ ,
\end{equation}
where $\hat{a}$ represents the estimated person location, and $\mathbf{1}()$ is an indicator function that equals 1 when the condition is satisfied and 0 otherwise. We then take the midpoint $(\hat{x}, \hat{y})$ of the bottom edge of the anchor box $\hat{a}$ as the person's coordinate in pixel space.
Here, we want to emphasize that our system is not limited to a specific OD method; users can choose the method based on their device capabilities and specific requirements. Current OD methods can primarily be categorized into two groups: model-based methods and deep learning methods. Model-based methods have the lowest computational cost but often require optimal environmental conditions to perform well, such as good lighting, a solid background color, and individuals dressed in bright colors. Unfortunately, these conditions may not always be met in practice. In contrast, deep learning methods offer better robustness without strict requirements on experimental conditions. We consider YOLO \cite{YOLO}, an early deep learning method, to be an excellent choice due to its high efficiency, speed, and precision. 

To convert the pixel coordinates to physical coordinates, we use the `getPerspectiveTransform' function from the OpenCV library \cite{OpenCV}. This function requires the coordinates of four non-collinear anchor points (reference points) in both pixel and physical spaces to calculate the transformation matrix $\mathbf{T}$ by solving:
\begin{equation}
[X'_1, X'_2, X'_3, X'_4] = \mathbf{T} \cdot [X_1, X_2, X_3, X_4] \ ,
\label{transfer}
\end{equation}
where $X_i$ and $X'_i$ represent the homogeneous coordinates of the four anchor points in pixel and physical spaces, respectively, and $\mathbf{T}$ is the transformation matrix. 
We then calculate the real-world person location coordinate $(\hat{x}', \hat{y}')$ by: 
\begin{equation}
(\hat{x}', \hat{y}') = \mathbf{T} \cdot (\hat{x}, \hat{y}) \ .
\end{equation}
The selection of the reference points is arbitrary as long as any three of them are not collinear. This can be achieved through manual marking or by using ruler apps available on mobile phones.



\subsection{Signal Collection and Completion}

During the image collection process, a Wi-Fi RX is simultaneously used for signal sensing. For instance, in Fig. \ref{fig:workflow}, an ESP32-S3 pings a Wi-Fi router, which serves as the transmitter (TX), with a frequency of $f_2$ and receives the corresponding CSI sequence. However, due to various factors such as device errors and environmental noise, packet loss frequently occurs. Research \cite{CSI-BERT} has demonstrated that this loss can significantly degrade network performance. To address this issue, we first identify the occurrences of packet loss. The RX receives a signal sequence $s = [s_1, s_2, \ldots, s_n]$ with corresponding timestamps $t = [t_1, t_2, \ldots, t_n]$. Ideally, the time gap between two consecutive timestamps should be approximately $\frac{1}{f_2}$. Therefore, the number of lost packets between two frames can be calculated as:
\begin{equation}
k=\max \left\{\text{round}\left( (t_{i+1} - t_i) f_2 - 1\right), 0\right\} \ .
\end{equation}
We can then fill in the missing data using neural networks or interpolation methods. Due to page limitations, further details can be found in \cite{CSI-BERT}.


\subsection{Modality Alignment}
Finally, we need to align the signal samples with the person coordinate labels. In practice, both the camera and Wi-Fi RX are synchronized to the same device, such as a personal computer (PC), for data collection, ensuring that the timestamps for the received images and signals are unified. This allows us to achieve alignment based on these timestamps. For a signal frame with timestamp $t_i$, we select the nearest image as its correspondence: 
\begin{equation}
m= arg \min \{|t_i - T|\} \ ,
\end{equation}
where $T = [T_1, T_2, \ldots, T_n]$ is the timestamp sequence of the images. In this way, the person coordinates in image $m$ are designated as the ground truth for signal frame $i$.

\begin{figure*}[t!]
\centering 
\subfloat{\includegraphics[width=0.4\textwidth]{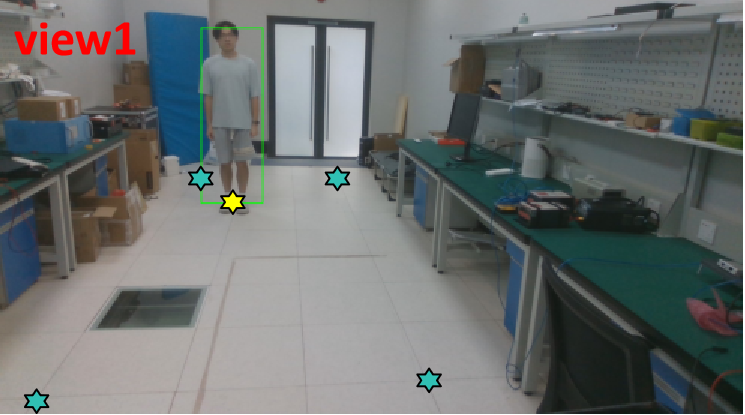}} 
\subfloat{\includegraphics[width=0.4\textwidth]{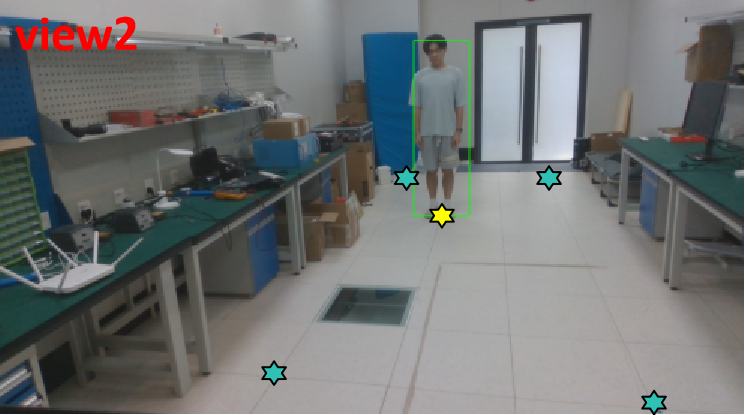}} \\
\subfloat{\includegraphics[width=0.4\textwidth]{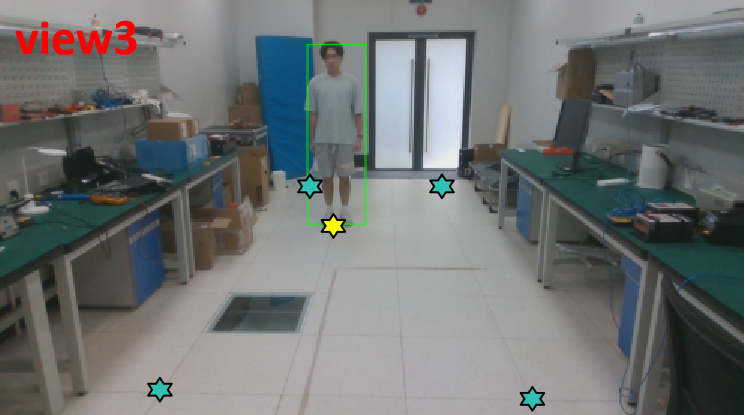}}
\subfloat{\includegraphics[width=0.4\textwidth]{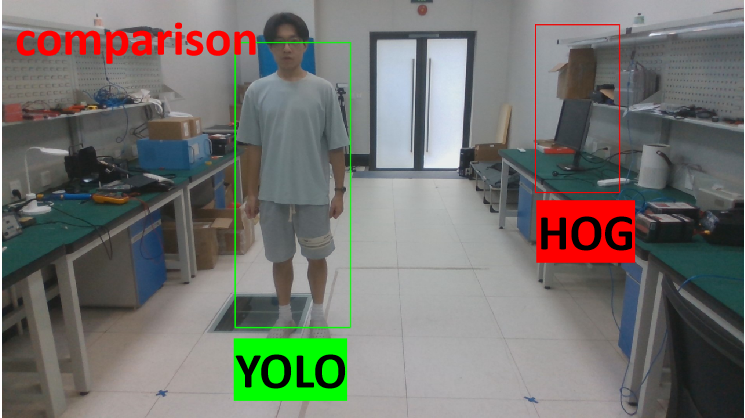}}
\caption{ Object Detection Results: The left three images illustrate the output of YOLO \cite{YOLO} from three different perspectives. The yellow stars indicate the positions of detected individuals, while the green stars represent anchor points. The rightmost image displays the results from YOLO, HOG \cite{dalal2005histograms}, and Haar Cascade \cite{viola2001rapid}; the green box corresponds to YOLO's results, the red box to HOG's results, and there is no box for Haar Cascade as it failed to detect any objects.}
\label{fig:view}
\end{figure*}


\section{Experiment}
\subsection{Dataset Description}
According to our LoFi method, we collect a Wi-Fi tracking dataset in a laboratory using a webcam and an ESP32-S3 with a single antenna. The ESP32-S3 receives the CSI and RSSI embedded in Ping Replies from a home Wi-Fi router, with a setting of 100Hz sampling rate and 2.4 GHz carrier frequency. The webcam is used to capture images at the same time as the ESP32-S3 with a sampling rate of 26Hz. During data collection, we recruited 7 volunteers and asked them to move arbitrarily (e.g., walk forward, walk backward, and stand) in a given region ($1.8m \times 4.8m$) for 5 minutes, respectively. We selected the four vertices of this rectangle as our reference points. For privacy considerations, we do not provide the raw images. Instead, we have already processed the CSI and image data using our LoFi method and correlated the individual coordinates to each CSI frame. Unlike some previous datasets, we do not split the CSI sequence. As a result, users can split the data using any sliding window. As each CSI frame has its corresponding individual coordinates, the effective data amount of our dataset can reach approximately 210 thousand.

To illustrate the precision of our label generator method, we tested the label error against ground truth, where we had two volunteers stand at 10 predefined points. We test the error by setting the camera in three different views, as shown in Fig. \ref{fig:view}. The mean errors are 17.44cm, 12.13cm, and 11.82cm, respectively. This error is acceptable because we consider each person as a single point, even though a person actually occupy a given size in space.
Here, we also compare two additional model-based algorithms, Haar Cascade \cite{viola2001rapid} and Histogram of Oriented Gradients (HOG) \cite{dalal2005histograms}. We observed that Haar Cascade sometimes experiences detection failures, while HOG occasionally detects incorrect objects, making both methods challenging to utilize. As shown in the rightmost subfigure of Fig. \ref{fig:view}, the green box generated by YOLO indicates a correct detection, the red box generated by HOG indicates an incorrect detection, and there is no box for Haar Cascade, indicating that it detected nothing.


\subsection{Benchmark Methods}
\begin{figure*}[t!]
\centering 
\subfloat[Convolution-based]{\includegraphics[width=0.35\textwidth]{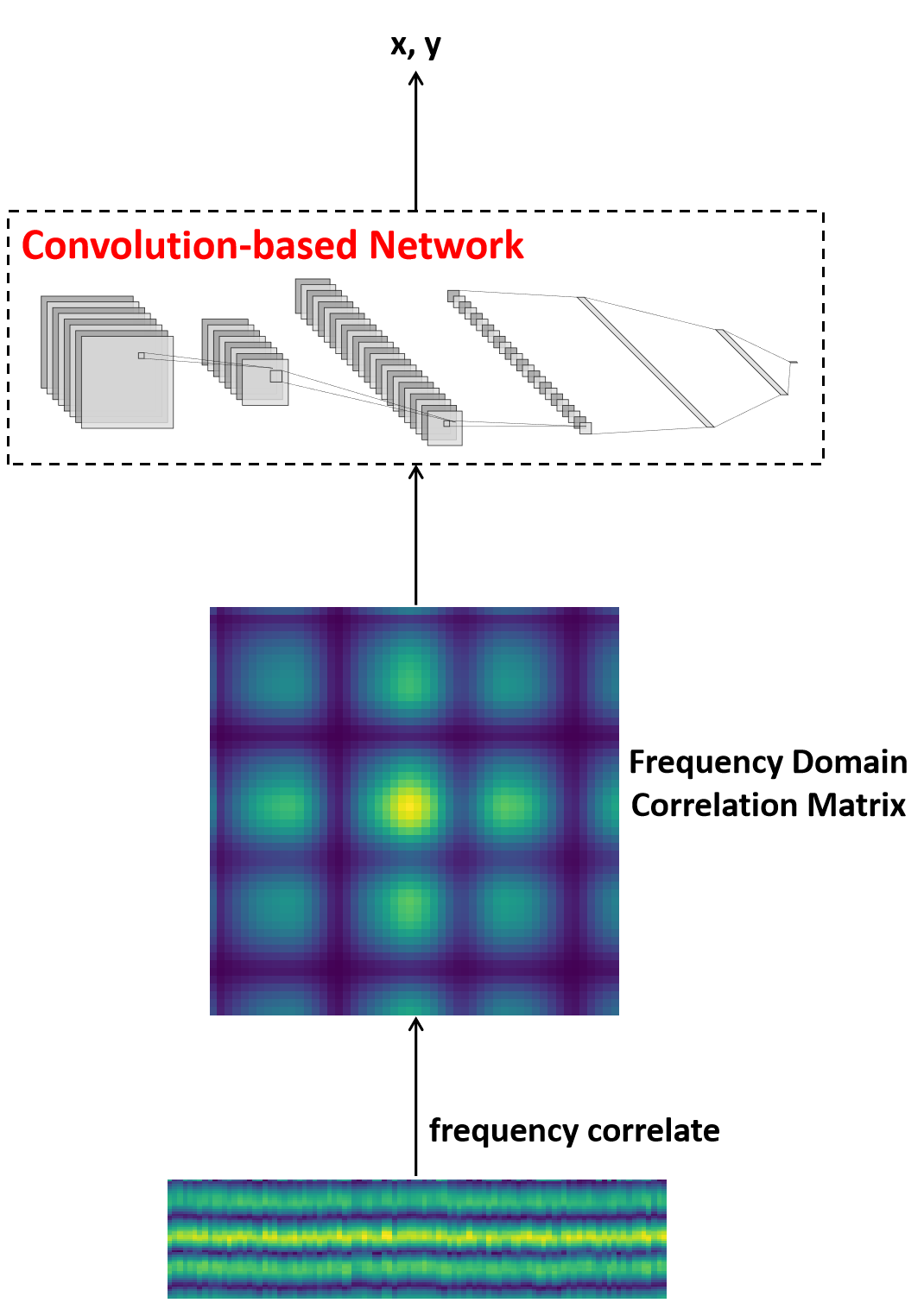}} 
\hspace{60pt} 
\subfloat[Series-based]{\includegraphics[width=0.35\textwidth]{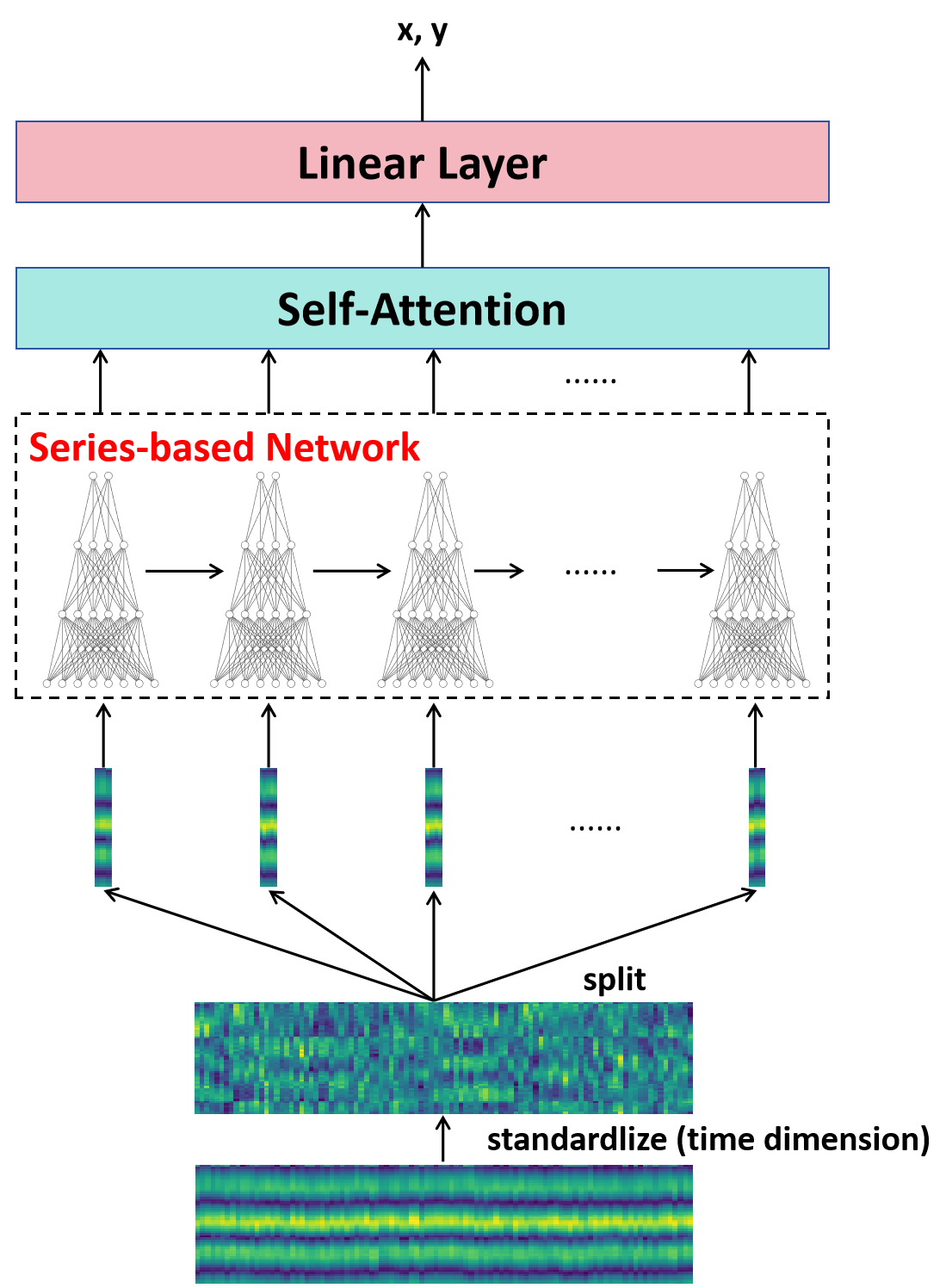}} \\
\caption{Architecture of Two Types of Benchmark Networks}
\label{fig:network}
\end{figure*}

In this section, we provide two common types of benchmark data-driven methods in Wi-Fi sensing: the convolution-based method and the series-based method, as shown in Fig. \ref{fig:network}.

For the convolution-based network, we first calculate the frequency domain correlation matrix as $F = M^T \cdot M$,
where $M \in R^{t,s}$ is the amplitude spectrum of CSI, $t$ is the time length, $s$ is the number of subcarriers, $F$ represents the correlation between different subcarriers. Then, the convolution network can extract features and predict the person coordinates based on this information.
And for the series-based network, we first standardize each subcarrier in the time dimension, following the approach in \cite{CSI-BERT}. Then, the standardized $M$ can be viewed as a time series and input to the series-based network. Finally, we use a self-attention layer and a linear layer to predict the person coordinates based on the features extracted by the series-based network.

\begin{table*}[t!]
\caption{Experiment Result: The bold value indicates the best result in each row.}
    \centering
        \begin{tabular}{|c|c|c|c|c|c|c|}
        \hline
        \multirow{2}{*}{\textbf{\diagbox{Metric}{Methods}}} & \multicolumn{2}{c|}{\textbf{Convolution-based Methods}} & \multicolumn{4}{c|}{\textbf{Series-based Methods}}\\
        \cline{2-7}
         & \textbf{CNN} & \textbf{ResNet \cite{Resnet}}  & \textbf{RNN \cite{RNN}} & \textbf{GRU \cite{GRU}} & \textbf{LSTM \cite{LSTM}} & \textbf{CSI-BERT \cite{CSI-BERT}} \\
        \hline 
         \textbf{Model Size} & \textbf{23K}  &  11M &  33K &100K &   133K &    2M \\
        \hline 
        \textbf{GPU Occupation (GB)} &  0.65  &  0.76  & \textbf{0.45}  &  0.49 &  0.49 &  0.88 \\
        \hline \hline
        \textbf{Error Mean (m)} & 0.8745 & \textbf{0.5830} & 0.8705 & 0.9413 & 0.8643 & 0.6991 \\
        \hline 
        \textbf{Error Standard Deviation (m)} & 0.3177 & 0.3475 & 0.2802 & 0.3217 & \textbf{0.2475} & 0.3063 \\
        \hline 
        \textbf{Classification Accuracy (6 classes)} & 31.99\% & 55.50\% & 53.29\% & 49.41\% & 53.50\% & \textbf{60.07\%} \\
        \hline 
        \textbf{Classification Accuracy (4 classes)} & 42.03\% & \textbf{62.54\%} & 61.92\% & 56.00\% & 62.15\% & 61.93\% \\
        \hline 
        \textbf{Classification Accuracy (2 classes)} & 62.84\% & 82.98\% & \textbf{84.47\%} & 73.56\% & 73.98\% &  75.63\% \\
        \hline 
        \end{tabular}
\label{tab:exp}
\end{table*}

\begin{figure}[htbp]
\centering 
\includegraphics[width=0.5\textwidth]{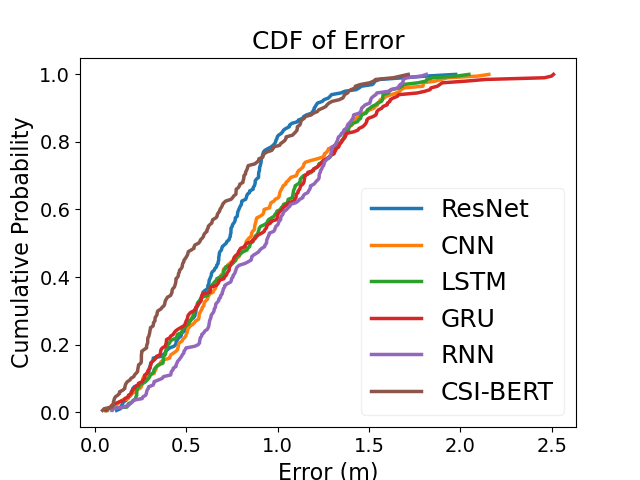}
\caption{CDF of Error for Benchmark Methods}
\label{fig:cdf}
\end{figure}

\subsection{Numerical Results}



In this section, we construct a series of benchmark networks and evaluate their performance on our collected dataset. 
For convolutional networks, we utilize a four-layer CNN and an 18-layer ResNet \cite{Resnet}. For sequence-based networks, we employ the four-layer RNN \cite{RNN}, GRU \cite{GRU}, LSTM \cite{LSTM}, and CSI-BERT \cite{CSI-BERT}. 
We train these models using the Adam optimizer with an initial learning rate of 0.001 for 200 epochs with batch size of 64, conducted on an Nvidia RTX 3090Ti GPU.
As shown in Table \ref{tab:exp}, we first evaluate the mean localization error and standard deviation. We observe that all methods exhibit excellent performance, with average errors lower than 1m. The ResNet and CSI-BERT achieve the lowest errors, likely due to their larger model sizes, which capture more robust features. Following the approach of many works \cite{SiFi}, we split the movement space into different regions and use the networks to predict the region in which a person is located. Here, ResNet and RNN show better performance in coarse-grained classification, while CSI-BERT demonstrates superior results in fine-grained classification. We then illustrate the Cumulative Distribution Function (CDF) of the errors in Fig. \ref{fig:cdf}, revealing that ResNet and CSI-BERT significantly outperform the others, achieving localization errors lower than 1 meter for approximately 80\% of the samples, whereas the other methods only reach about 60\%. Additionally, most previous Wi-Fi localization studies require multiple RX-TX pairs or multiple antennas. Our experiments reveal the potential of using deep learning methods for Wi-Fi localization with a single RX-TX pair and a single antenna.

\section{Conclusion}

In this paper, we propose LoFi, a vision-aided label generation method for Wi-Fi localization and tracking. Our approach offers a simple and cost-effective solution for extracting individual coordinates from 2D images, which can be captured using any camera device. This method provides a practical technique for data collection, facilitating the training or rapid fine-tuning of relevant models. Additionally, it can collect more diverse data without any restriction on people movement, which will aid the development of data-driven methods in this field. Furthermore, we have published a dataset using our method, which can be utilized in research on Wi-Fi localization and tracking, people identification, and cross-domain Wi-Fi sensing.


\section*{Acknowledgment}
This work was supported in part by National Natural Science Foundation of China (Grant No. 62371313), in part by ShenzhenHong Kong-Macau Technology Research Programme (Type C) (Grant No. SGDX20230821091559018), in part by the Shenzhen Science and Technology Program (Grant No. JCYJ20241202124934046), in part by Guangdong Young Talent Research Project (Grant No. 2023TQ07A708).

\bibliographystyle{ieeetr}
\bibliography{ref}


\end{document}